\documentclass[conference]{IEEEtran}
\usepackage{cite}
\usepackage{graphicx}
\graphicspath{ {images/} }

\usepackage{subfig}
\usepackage{booktabs}
\usepackage{amsfonts}
\usepackage{amsmath}
\usepackage{multirow}
\usepackage[style=base]{caption}

\begin{document}

\pdfinfo{/Title (STDP Learning of Image Features with Spiking Neural Networks) /Author (Daniel J. Saunders, Hava T. Siegelmann, Robert Kozma, Miklos Ruszinko)}

\title{STDP Learning of Image Patches with \\Convolutional Spiking Neural Networks}

\author{
\IEEEauthorblockN{Daniel J. Saunders\IEEEauthorrefmark{1}, Hava T. Siegelmann\IEEEauthorrefmark{2}, Robert Kozma\IEEEauthorrefmark{3}} 
\IEEEauthorblockA{
College of Information and Computer Sciences \\
University of Massachusetts Amherst \\
140 Governors Drive, Amherst, MA 01003, USA \\
email: \{djsaunde\IEEEauthorrefmark{1}, hava\IEEEauthorrefmark{2}, rkozma\IEEEauthorrefmark{3}\}@cs.umass.edu
}
\and
\IEEEauthorblockN{Mikl\'{o}s Ruszink\'{o}}
\IEEEauthorblockA{
	Alfr\'{e}d R\'{e}nyi Institute of Mathematics \\
Hungarian Academy of Sciences \\
13-15 Re\'{a}ltanoda utca, H-1053 Budapest, Hungary \\
email: ruszinko.miklos@renyi.mta.hu
}}

\maketitle

\begin{abstract}

Spiking neural networks are motivated from principles of neural systems and may possess unexplored advantages in the context of machine learning. A class of \textit{convolutional spiking neural networks} is introduced, trained to detect image features with an unsupervised, competitive learning mechanism. Image features can be shared within subpopulations of neurons, or each may evolve independently to capture different features in different regions of input space. We analyze the time and memory requirements of learning with and operating such networks. The MNIST dataset is used as an experimental testbed, and comparisons are made between the performance and convergence speed of a baseline spiking neural network.

\end{abstract}


\section{Introduction}
\label{intro}

Deep learning has been shown to be wildly successful across a wide range of machine learning tasks. Without large labeled data and plentiful computational resources for training, however, this approach may not be appropriate \cite{y._lecun_deep_2015}. Unsupervised learning, algorithms that learn a representation of a dataset without a supervision signal, is a solution to the former problem. Although several unsupervised learning approaches perform well in certain settings, a robust approach to unsupervised learning in general has yet to be discovered. Further, modern approaches to unsupervised often still require massive computation and unlabeled data in order to achieve respectable performance. 

Real neuronal systems need only little data to recognize recurring patterns and are able to dynamically manage energy resources while solving complex problems \cite{lee_neural_2015}. Motivated first from biological principles \cite{maass_lower_1996}, further development of \textit{spiking neural networks} (SNNs) should take advantage of these properties, and may motivate the development of general and flexible machine learning. While possessing several biologically-plausible elements, it has been shown that certain SNNs inhabit a strictly more powerful class of computational models \cite{maass_networks_1997} than that of traditional neural networks. Moreover, SNN training and evaluation has the potential to be \textit{massively parallelized} while scaling modestly with the size of input data and number of neurons used.

Building on SNNs studied in \cite{p._u._diehl_unsupervised_2015}, architectural modifications are introduced to improve the capacity of spiking neural networks to represent image data. Inspired by the success of convolutional neural networks on many image processing tasks, we propose \textit{convolutional spiking neural networks} (C-SNNs), a class of networks which learn features of grid-like data in an unsupervised fashion. Small \textit{patches} of neurons may share parameters to improve training efficiency, or may evolve independently to learn separate features. New inhibitory connectivity is introduced in which neurons in different sub-populations compete to represent isolated regions of input space, enabling fast convergence to a robust data representation and respectable classification accuracy. 

We show examples of C-SNN learned representations, and present classification results based on network filters and spiking activity on the MNIST handwritten digit dataset \cite{y._lecun_gradient-based_1998}.

\section{Related Work}
\label{related}

This paper builds on the work of Diehl and Cook \cite{p._u._diehl_unsupervised_2015} where a spiking neural network (SNN) is used to classify the MNIST handwritten digits after learning network weights without supervision and with several \textit{spike-timing-dependent plasticity} (STDP) rules. Classification performance of their networks increases with the number of spiking neurons used, ultimately achieving approximately 95\% accuracy using networks of 6,400 excitatory and inhibitory neurons.

A number of other SNNs trained with STDP are used to classify image data \cite{liu_fast_2017}, \cite{kheradpisheh_stdp-based_2016}. The former uses Gabor filter features as a pre-processing input to their network, uses the rank-order coding scheme for input spikes, and classifies data with the winner-take-all strategy on the output layer. The latter is comprised of a difference of Gaussians pre-processing step, followed by convolutional and pooling layers, and whose output is trained on a linear SVM to perform classification. Other systems use spiking neurons, but are trained with supervision; e.g. \cite{diehl_fast-classifying_2015}, which was first trained as a deep neural network using back-propagation and later transferred to a spiking neural network without much loss in performance.

\section{Methods}
\label{Methods}

\subsection{LIF neuron with synaptic conductances}

One step towards a more biologically realistic learning system is to include a more powerful neural unit. The basic computational operations of a deep neural network (DNN) do not incorporate time, and unlike the all-or-nothing action potential of the biological neuron, they communicate by sending precise floating-point numbers downstream to the next layer of processing units. Moreover, the standard neuron that is used in DNNs and other traditional neural networks is synchronous, without memory of previous actions, in contrast to spiking neurons which are asynchronous and integrate time in their internal operation.

A simple choice of a biologically-inspired unit of computation is the leaky integrate-and-fire (LIF) spiking neuron \cite{w._gerstner_spiking_2002}. We consider a modified version of the LIF neuron, which implements a \textit{homeostasis} mechanism to balance overall network activity. LIF neurons naturally incorporate time in their operation, dynamically changing membrane potential and spiking threshold, yet are computationally simple enough to scale to large networks trained on many data examples. In our models, we use a network of LIF neurons, both \textit{excitatory} and \textit{inhibitory}, and additionally model \textit{synaptic conductances}. Their membrane voltage $v$ is given by

\begin{gather}
\tau \frac{\textrm{d}v}{\textrm{dt}} = (v_{rest} - v) + g_e (E_{exc} - v) + g_i (E_{inh} - v),
\end{gather}

where $v_{rest}$ is the resting membrane potential, $E_{exc}$ and $E_{inh}$ are the equilibrium potentials of excitatory and inhibitory synapses, and $g_e$ and $g_i$ are the conductances of excitatory and inhibitory synapses. The time constant $\tau$ is chosen to be an order of magnitude larger for excitatory neurons than for inhibitory neurons. When a neuron's membrane potential exceeds its membrane threshold $v_{\textrm{threshold}}$, it fires an action potential (spike), increases $v_{\textrm{threshold}}$ and resets back to $v_{\textrm{reset}}$. The neuron then undergoes a short refractory period (approximately 5ms), during which it cannot fire any spikes.

\begin{figure}
	\centering
    \captionsetup{justification=centering}
	\includegraphics[width=0.475\textwidth, height=5cm]{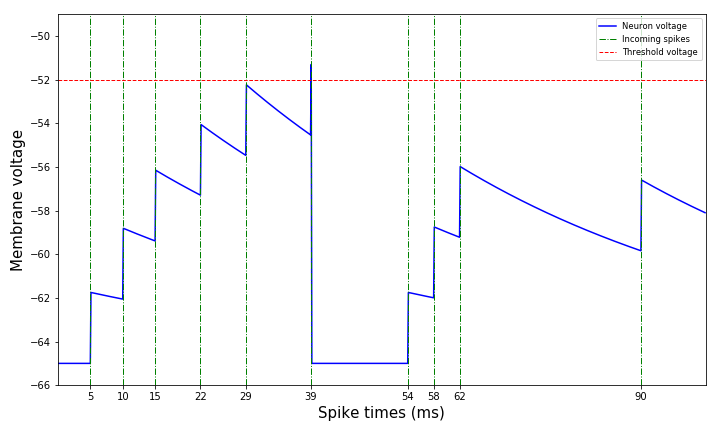}
	\caption{Membrane voltage is increased as neurons integrate incoming spikes and is otherwise decaying. Once the voltage exceeds $v_\textrm{threshold}$, it is reset to $v_\textrm{reset}$.}
	\label{fig:voltage}
\end{figure}

Figure \ref{fig:voltage} shows a recording of a single neuron's voltage as it integrates incoming action potentials as an instantaneous voltage change. Otherwise, voltage is decaying exponentially to $v_\textrm{rest} = -65$mV. At time $t = 39$, the neuron's membrane voltage exceeds $v_\textrm{threshold} = -52$mV and resets to $v_\textrm{reset} = -65$mV.

Synapses are modeled by conductance changes: a synapse increases its conductance when a presynaptic action potential arrives at the synapse by its synaptic weight $w$. Otherwise, the conductance is decaying exponentially. The dynamics of the synaptic conductance are given by

\begin{gather}
\tau_{g_e} \frac{\textrm{d}g_e}{\textrm{dt}} = -g_e, \hspace*{0.15cm} \tau_{g_i} \frac{\textrm{d}g_i}{\textrm{dt}} = -g_i.
\end{gather}

\subsection{Spike-timing-dependent plasticity}

STDP is used \cite{bi_synaptic_1998} to modify the weights of synapses connecting certain neuron groups. We call the neuron from which a synapse projects \textit{presynaptic}, and the one to which it connects \textit{postsynaptic}. For the sake of computational efficiency, we use online STDP, where values $a_{\textrm{pre}}$ and $a_{\textrm{post}}$ (\textit{synaptic traces}) are recorded for each synapse, a simple decaying memory of recent spiking history. Each time a presynaptic (postsynaptic) spike occurs at the synapse, $a_{\textrm{pre}}$ ($a_{\textrm{post}}$) is set to 1; otherwise, it decays exponentially to zero with a time course chosen in a biologically plausible range. When a spike arrives at the synapse, the weight change $\Delta w$ is calculated using a STDP learning rule. The rule we use in our simulations is given by

\begin{gather}
\Delta w = 
\begin{cases}
\eta_{\textrm{post}} a_{\textrm{pre}} (w_{\textrm{max}} - w) & \text{on postsynaptic spike} \\
- \eta_{\textrm{pre}} a_{\textrm{post}} w & \text{on presynaptic spike}
\end{cases}
\end{gather}

\begin{figure}[h]
	\centering
    \captionsetup{justification=centering}
	\frame{\includegraphics[width=0.45\textwidth, height=5.75cm]{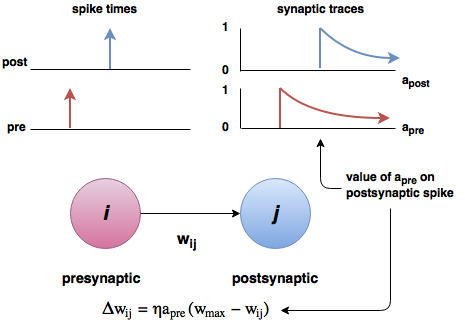}}
	\caption{The magnitude and sign of an STDP update is determined by the ordering and relative timing of pre- and post-synaptic neurons.}
	\label{fig:STDP}
\end{figure}

The terms $\eta_{\textrm{post}}, \eta_{\textrm{pre}}$ denote learning rates, and $w_{\textrm{max}}$ is the maximum allowed synaptic weight. There are many options for STDP update rules \cite{p._u._diehl_unsupervised_2015}, but we chose this form of update due to its computational efficiency and the observation that many Hebbian-like rules produced similar learning behavior in our SNNs. Our STDP rule is illustrated in Figure \ref{fig:STDP}.

All simulations are accomplished using the BRIAN spiking neural networks simulator \cite{d._f._m._goodman_brian_2009}.

\subsection{Input encoding and simulation}
\label{ssec:input}

MNIST examples are encoded in a population of Poisson spiking neurons \cite{w._gerstner_spiking_2002}. Each \textit{input neuron} is given an average firing rate $\lambda_i$ proportional to the intensity of the $i$-th input pixel. In both, baseline SNNs and C-SNNs, this same \textit{input layer} is used to present the input data in an inherently noisy and time-varying fashion. This is in contrast to a typical machine learning approach, where the input data is immediately consumed at the first step of a learning or inference step. However, it is not uncommon to corrupt input data with noise in order to enforce a more robust data representation during training.

A single training iteration lasts for 350ms of simulation time (equations are evaluated every 0.5ms), and networks are run without input for 150ms between each example to ``relax'' back to equilibrium before the next iteration begins.

\subsection{SNN architecture}

To compare with the C-SNN, we consider a class of SNNs \cite{p._u._diehl_unsupervised_2015} consisting of a collection of three populations of neurons, or \textit{layers}. The \textit{input layer} is as described in Section \ref{ssec:input}, with a number of input neurons equal to the dimensionality of the input space. 

The \textit{excitatory layer} is composed of an arbitrary number of excitatory neurons, each of which receive STDP-modifiable synapses from the input layer. Each of these neurons connect to a unique neuron in the \textit{inhibitory layer}, which connects back to all excitatory neurons, except for the neuron from which it receives its connection. The latter two connections are fixed; i.e., not modified by STDP during the training phase.

The task is to learn a representation of the dataset on the synapses connecting the input to the excitatory layer. In this case, the connectivity pattern between excitatory and inhibitory layers creates competition between neurons in the excitatory layer in which, after an initial transient at the start of each iteration, a firing excitatory neuron typically causes all other excitatory neurons to fall quiescent. 

\begin{figure}[ht]
  \centering
  \captionsetup{justification=centering}
  \includegraphics[width=0.475\textwidth, height=6cm]{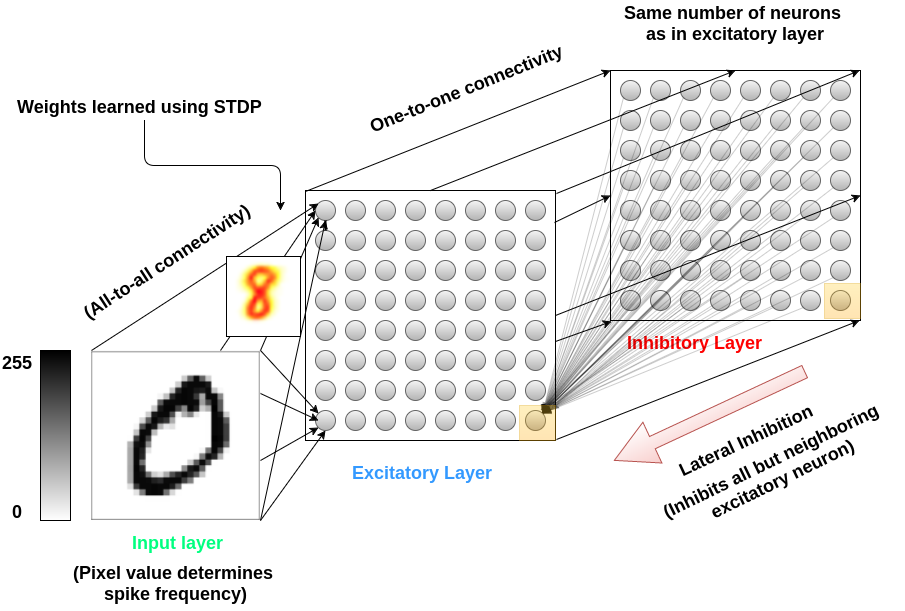}
  \caption{A depiction of the baseline spiking neural network architecture \cite{p._u._diehl_unsupervised_2015}. Weights from the input to excitatory layer are adapted to capture typical input examples.}
  \label{fig:SNN}
\end{figure}

\begin{figure}[ht]
  \centering
  \captionsetup{justification=centering}
  \includegraphics[width=0.475\textwidth, height=9cm]{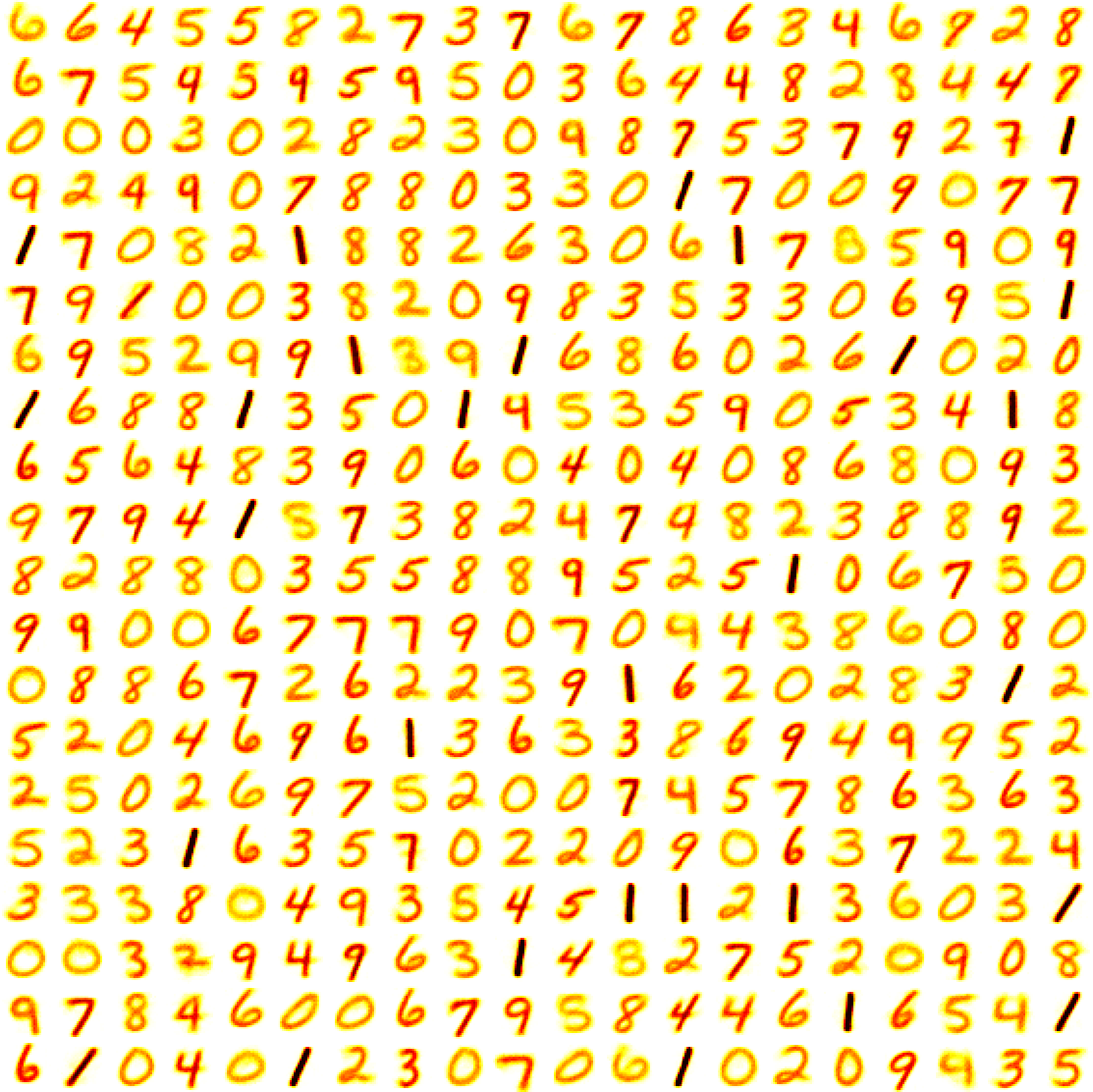}
  \caption{An example set of 400 learned filters from the baseline SNN model, specialized to different classes of data as a result of a competitive inhibition mechanism.}
  \label{fig:Learned filters}
\end{figure}

This behavior allows individual neurons to learn unique filters: the single most-spiked neuron on each iteration updates its synapse weights the most to match the current input digit, per the chosen STDP rule. Increasing the number of excitatory and inhibitory neurons has the effect of allowing a SNN to \textit{memorize} more examples from the training data and recognize similar patterns during the test phase. The SNN network architecture is illustrated in Figure \ref{fig:SNN}, and an example set of \textit{filters} (for an SNN with 400 excitatory and inhibitory neurons) is shown in Figure \ref{fig:Learned filters}.

\subsection{C-SNN architecture}

As with the SNN architecture, C-SNNs are organized into input, excitatory, and inhibitory neuron populations. We introduce parameters $k$, giving a convolution kernel size $k \times k$, $s$, horizontal and vertical convolution stride, and $n_{\textrm{patches}}$, giving the number of convolution \textit{patches} to use. A convolution patch is a sub-population of excitatory neurons (coupled with equally sized inhibitory populations), the size of which is determined by parameters $k$ and $s$. A window of size $k \times k$ ``slides'' across the input space with stride $s$. Each portion of the input space, characterized by where the window falls, is connected with STDP-modifiable synapses to a single neuron in each of the $n_\textrm{patches}$ excitatory sub-populations.

Similar to the SNN connectivity, neurons in the excitatory sub-populations connect one-to-one with equally-sized inhibitory sub-populations; however, inhibitory neurons connect only to excitatory neurons in other sub-populations which share the same \textit{visual field}; i.e., receive connections from the same region of input space.

\begin{figure}[ht]
  \centering
  \captionsetup{justification=centering}
  \includegraphics[width=0.475\textwidth, height=5.75cm]{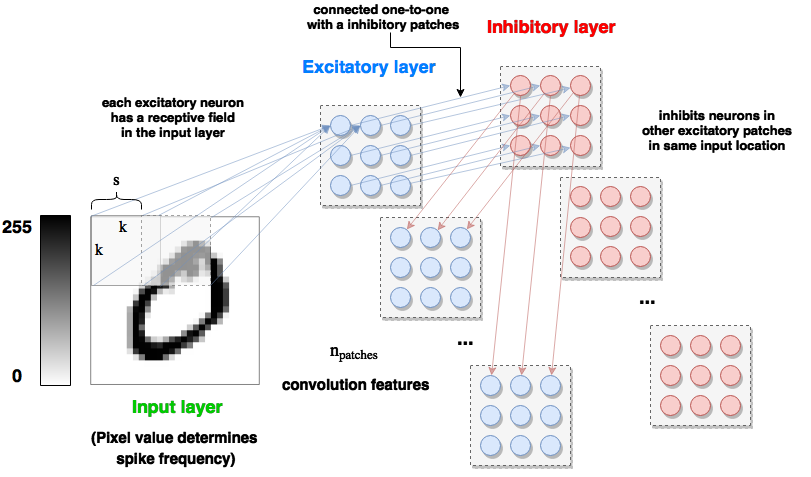}
  \caption{A depiction of the proposed convolutional spiking neural network architecture. }
  \label{fig:C-SNN}
\end{figure}

\begin{figure}[ht]
  \centering
  \captionsetup{justification=centering}
  \includegraphics[width=0.475\textwidth, height=4.5cm]{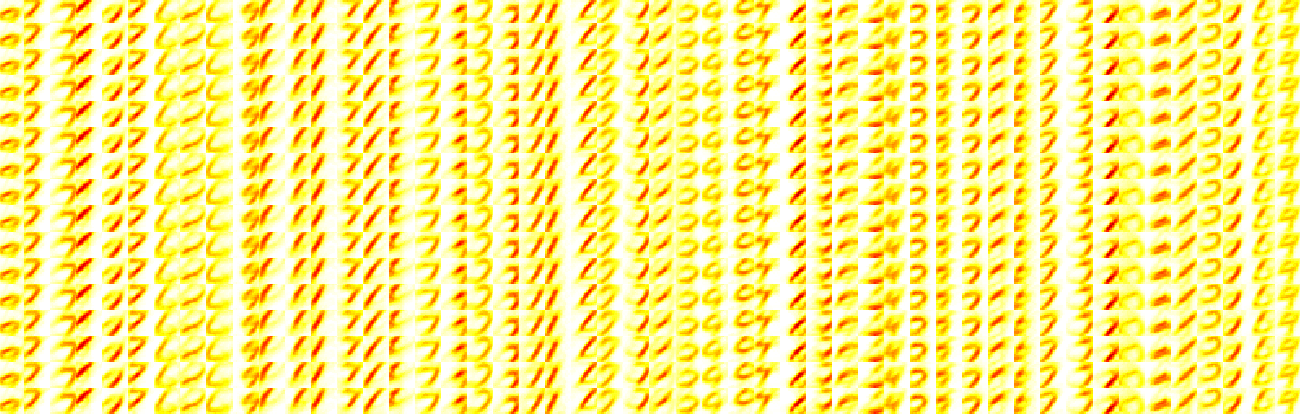}
  \caption{Example shared weights: Each column corresponds to a unique feature learned by a single excitatory patch.}
  \label{fig:shared_weights}
\end{figure}

\begin{figure}[ht]
  \centering
  \captionsetup{justification=centering}
  \includegraphics[width=0.475\textwidth, height=4.5cm]{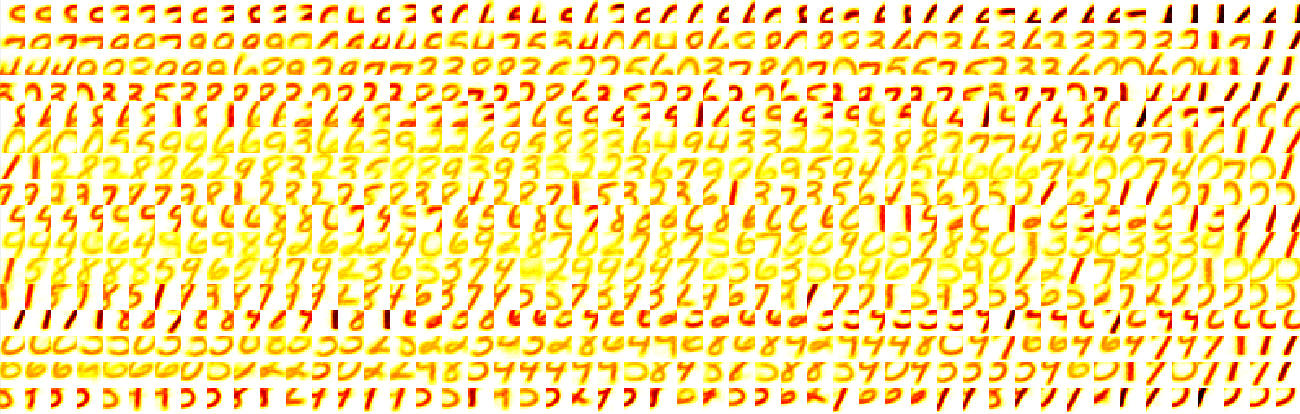}
  \caption{Example unshared weights: Rows corresponds to unique locations in the input space, ranging from upper left to lower right corners.}
  \label{fig:unshared_weights}
\end{figure}

A schematic of this architecture is shown in Figure \ref{fig:C-SNN}. Example learned features for both \textit{shared} and \textit{un-shared weights} are shown in Figures \ref{fig:shared_weights} and \ref{fig:unshared_weights}. The columns index different excitatory neuron sub-populations, and rows index locations in the input space. In both cases, we set $n_\textrm{patches} = 50$, $k = 16$, and $s = 4$, which produces $n_\textrm{patch} = 16$ neurons per excitatory and inhibitory patch.

In Figure \ref{fig:unshared_weights}, each neuron has learned a unique \textit{filter} or feature which may only appear in their visual field. Those neurons with visual fields near the edge of the input space tend to produce highly converged filters, while those near the center tend to be less confident. We argue that this is due to the fact that most inputs lie near the center of the input space, while relatively few examples extend to the borders. On the other hand, Figure \ref{fig:shared_weights} demonstrates how, with shared weights, C-SNNs may learn a set of features which are not necessarily dependent on any particular region of input space. Such features tend to be invariant to locations in the input while maintaining an interesting summarization of the properties of the data.

\begin{figure}[ht]
  \centering
  \includegraphics[width=0.45\textwidth, height=5cm]{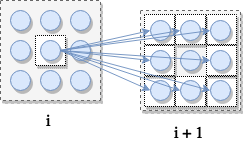}
  \caption{Between-patch lattice connectivity}
  \label{fig:lattice_connectivity}
\end{figure}

Excitatory sub-populations, or \textit{patches}, may be connected via STDP-modifiable synapses in several ways. We consider \textit{between-patch connectivity} in which neurons in one excitatory patch are connected to neurons in another patch whose visual field borders that of the former neuron's. The neighborhood is described by a number $m$ of a neuron's neuron spatial neighbors, which creates a lattice-like connectivity pattern between patches. The sub-populations which are connected may also be varied; one subpopulation may connect to all others, or only to a neighbor. A schematic of this connectivity for patches of size $3 \times 3$ using a lattice neighborhood of size 8 is depicted in Figure \ref{fig:lattice_connectivity}.

\subsection{Evaluating learned representations}

Although LM-SNNs are trained in an unsupervised manner, we may want to evaluate the quality of the representations they learn. The dataset representation is encoded in the learned weights of synapses connecting the input and excitatory layers. We use the activity of the neurons in the excitatory layer with their filter weights held fixed to (1) say what it means to \textit{represent} input data from a certain class, and (2) \textit{classify} new inputs based on historical network activity.

We perform a two-step procedure before the test phase to \textit{label} excitatory neurons with the input category they represent, and then \textit{classify} new data based on these labels and the spiking activity of the excitatory neurons on it, as in \cite{p._u._diehl_unsupervised_2015}. We call the first step \textit{neuron labeling} and the second \textit{voting}. We consider a single neuron labeling method, in which excitatory neurons are assigned a label (0-9) according to its highest class-wise averaged firing rate over all training examples. Several different methods of voting were developed and evaluated in the C-SNN framework:

\textit{all}: All excitatory neuron spikes are counted in the vote for the label of the input example. The data label with the highest average spiking rate is then used to classify the input.

\textit{most-spiked}: The label assignment of the neuron which emitted the most spikes is used to classify the current input example.

\textit{top percent}: Only the most-spiked $p$ percentile of neurons are permitted to cast a vote for the label of the input example. We chose a default value of $p = 10\%$, effectively preventing many low-activity (spurious) neurons from contributing to the classification of new data. Again, the label with the highest average spiking rate is chosen as the classification.

\textit{correlation clustering}: On each training iteration, a vector of the indices of the most-spiked neurons per excitatory patch is recorded, along with the true input label. In the test phase, these vectors are created again for each input example. The test vectors are compared against the training vectors, and the label of the training vector with the minimal distance from a test vector is used to classify the test example.

In Section \ref{sec:Results}, we evaluate networks with the above labeling and voting strategies.

\subsection{Computational complexity}

A potential advantage of the spiking neural networks approach is the amount of computation required for training and testing relative to large deep learning models. SNNs trained with spike-timing-dependent plasticity are trained in \textit{real time}; i.e., there is no need for both a forward and backward pass through the network, as with deep learning models trained with back-propagation \cite{Rumelhart:1995:BBT:201784.201785}. Instead, all synapse weights evolve independently, according to the relative timing of spikes emitting by their pre- and post-synaptic neurons. This \textit{local learning} rule is the key to training large networks while removing interdependence between learned parameters.

Let $m$ denote the number of excitatory and inhibitory neurons, $n$ the number of input neurons, and $k$ the number of convolution patches in a C-SNN. Weights from input to excitatory layers are continuously recorded, incurring a memory cost of $n m$. Membrane voltages for all excitatory, inhibitory neurons are updated on each timestep, requiring $2m$ calculations, and also update synaptic conductances for all inter-population connections, totaling $n m + m + (m - 1)^2$ updates per time step, which incur an equally-sized memory cost. Synaptic traces are recorded for input to excitatory connections; i.e., $n m$ values recording a fading memory of pre- and post-synaptic spike activity. Synapse weight updates are performed as needed: in the event of a spike from a pre- or post-synaptic neuron, we update the strength of the connecting synapse according to the value of the synaptic traces and the form of the STDP learning rule. Depending on the input data, both input and excitatory neurons spike at average rates $r_X, r_E$, respectively, requiring approximately $r_X m + r_E n$ weight updates per timestep. We estimated that $r_X \approx 4 \times 10^{-3}$ and $r_E \approx 2.75 \times 10^{-4}$ while training C-SNN networks of several sizes on the MNIST digit dataset for the first 1000 examples with a 0.5ms timestep. As the training phase progresses, fewer weight updates are made since network weights have converged to useful filters.

To simulate network dynamics and spiking activity for $T$ timesteps per example, the SNN and LM-SNN architectures require $T (3m + 2 nm + (m - 1)^2 + r_X m + r_E n) = \mathcal{O}(T (nm + m^2))$ operations, and $3nm + (m - 1)^2 + 3m = \mathcal{O}(nm + m^2)$ memory overhead per data item. We use a default of $T_1 = 700$ (350ms at 0.5ms timestep) for each example, and $T_2 = 300$ (150ms at 0.5ms timestep) for inter-example network relaxation periods. However, per-example simulation time $T_1$ can be reduced by appropriately adjusting simulation parameters to adapt filter weights at a quicker pace while maintaining stable network activity. Resetting of network state variables after each input example may allow the removal of the inter-example relaxation period.

Although the time and memory estimates are quadratic in the number of excitatory and inhibitory neurons, network simulation may tractably scale to larger or multi-layered SNNs by using parallel computation. Learning with STDP, there is no need to wait for a forward propagating signal to reach the network's output, or for an error signal back-propagation pass. Thus, network training may be massively parallelized to independently update synapse weights for all pairs of connected neurons. Improving network simulation to make better use of parallelization will drastically speed training and evaluation of spiking neural network architectures.

\section{Results}
\label{sec:Results}

We present classification accuracy results in Sections \ref{ssec:shared_weights} and \ref{ssec:unshared_weights}, and analyze the improved training convergence of C-SNNs to that of baseline SNNs \cite{p._u._diehl_unsupervised_2015} in \ref{ssec:convergence}. All networks are trained on 60,000 examples and evaluated on the 10,000 example test dataset. For every choice of model hyper-parameters, 10 independent trials are run, and the average test accuracy plus or minus a single standard deviation is reported.

Results for a baseline SNN \cite{p._u._diehl_unsupervised_2015} with various numbers of excitatory and inhibitory neurons are shown in Table \ref{fig:baseline} for comparison.

\begin{table}
\centering
\caption{baseline SNN \cite{p._u._diehl_unsupervised_2015}}
\label{fig:baseline}
\begin{tabular}{|c|c|}
\hline
$n_\textrm{neurons}$ & baseline SNN \\
\hline
100 & 80.71\% $\pm$ 1.66\% \\
225 & 85.25\% $\pm$ 1.48\% \\
400 & 88.74\% $\pm$ 0.38\% \\
625 & 91.27\% $\pm$ 0.29\% \\
900 & 92.63\% $\pm$ 0.28\% \\
\hline
\end{tabular}
\end{table}

\subsection{C-SNN: shared weights}
\label{ssec:shared_weights}

We present results for C-SNNs in which neurons in the same sub-populations share the same image feature. In Table \ref{shared_no_connectivity}, there is no connectivity between excitatory neuron patches; in Table \ref{shared_pairwise_connectivity}, there is lattice connectivity between each excitatory patch and those patches immediately neighboring it. We have fixed the number of sub-populations to 50 and the convolution stride length to 2. Reported accuracies are low across all settings of kernel window side length $k$ and evaluation strategies, yet some voting schemes are preferred depending on the choice of parameter $k$.

\begin{table}
\centering
\captionsetup{justification=centering}
\caption{shared weights, no \\connectivity, $n_\textrm{patches} = 50, s = 2$}
\label{shared_no_connectivity}
\scriptsize
\begin{tabular}{|c|c|c|c|c|}
\hline
$k$  & \textit{all}   & \textit{most-spiked} & \textit{top} \% & \textit{correlation}      \\
\hline
8  & 63.13 $\pm$ 1.54\% & 61.07 $\pm$ 1.67\% & 67.97 $\pm$ 1.45\% & 45.35 $\pm$ 1.87\% \\
10 & 66.86 $\pm$ 1.34\% & 68.99 $\pm$ 1.83\% & 72.50 $\pm$ 1.50\% & 54.53 $\pm$ 1.74\% \\
12 & 63.45 $\pm$ 1.61\% & 66.72 $\pm$ 1.48\% & 71.27 $\pm$ 1.42\% & 47.54 $\pm$ 1.68\% \\
14 & 64.31 $\pm$ 1.55\% & 66.79 $\pm$ 1.74\% & 69.19 $\pm$ 1.49\% & 63.31 $\pm$ 1.48\% \\
16 & 59.27 $\pm$ 1.29\% & 66.36 $\pm$ 1.58\% & 67.76 $\pm$ 1.71\% & 62.69 $\pm$ 1.63\% \\
18 & 56.15 $\pm$ 1.73\% & 66.35 $\pm$ 1.46\% & 67.40 $\pm$ 1.61\% & 63.49 $\pm$ 2.02\% \\
20 & 61.00 $\pm$ 1.19\% & 68.28 $\pm$ 1.74\% & 70.22 $\pm$ 1.55\% & 69.03 $\pm$ 1.33\% \\
22 & 63.84 $\pm$ 1.32\% & 66.90 $\pm$ 1.22\% & 69.22 $\pm$ 1.37\% & \textbf{72.90} $\pm$ 1.26\% \\
24 & 68.24 $\pm$ 1.09\% & 68.55 $\pm$ 0.98\% & 70.17 $\pm$ 1.05\% & 71.70 $\pm$ 1.17\% \\    
\hline
\end{tabular}
\end{table}

\begin{table}
\centering
\captionsetup{justification=centering}
\caption{shared weights, pairwise \\connectivity, $n_\textrm{patches} = 50, s = 2$}
\label{shared_pairwise_connectivity}
\scriptsize
\begin{tabular}{|c|c|c|c|c|}
\hline
$k$ & \textit{all} & \textit{most-spiked} & \textit{top} \% & \textit{correlation} \\
\hline
8   & 58.63 $\pm$ 1.69\% & 60.21 $\pm$ 1.77\% & 65.66 $\pm$ 1.07\% & 37.56 $\pm$ 1.42\% \\
10  & 62.92 $\pm$ 1.02\% & 63.86 $\pm$ 1.09\% & 68.33 $\pm$ 1.43\% & 41.67 $\pm$ 1.10\% \\
12  & 63.05 $\pm$ 0.95\% & 67.60 $\pm$ 1.07\% & 72.07 $\pm$ 1.58\% & 49.24 $\pm$ 1.36\% \\
14  & 65.13 $\pm$ 1.49\% & 69.74 $\pm$ 1.26\% & \textbf{72.24} $\pm$ 0.95\% & 62.20 $\pm$ 1.69\% \\
16  & 60.01 $\pm$ 1.03\% & 67.35 $\pm$ 0.89\% & 67.32 $\pm$ 1.35\% & 59.91 $\pm$ 1.44\% \\
18  & 57.21 $\pm$ 1.38\% & 66.19 $\pm$ 1.47\% & 69.85 $\pm$ 1.70\% & 63.35 $\pm$ 1.55\% \\
20  & 60.73 $\pm$ 1.45\% & 68.87 $\pm$ 1.06\% & 68.66 $\pm$ 1.08\% & 67.04 $\pm$ 1.59\% \\
22  & 65.93 $\pm$ 1.26\% & 66.91 $\pm$ 1.19\% & 67.32 $\pm$ 1.19\% & 70.27 $\pm$ 1.02\% \\
24  & 63.82 $\pm$ 1.54\% & 67.35 $\pm$ 1.32\% & 67.23 $\pm$ 1.36\% & 68.47 $\pm$ 1.50\% \\
\hline
\end{tabular}
\end{table}

\subsection{C-SNN: unshared weights}
\label{ssec:unshared_weights}

Test accuracy results for C-SNNs where neurons in the same sub-populations are allowed to learn independent features are shown in Tables \ref{unshared_no_connectivity} (no connectivity between excitatory patches) and \ref{unshared_pairwise_connectivity} (connectivity between neighboring excitatory patches). Again, the number of patches is fixed to 50, and the stride length is fixed to 2. Allowing each neuron to specialize to a unique filter gave an approximate 20\% increase in test accuracy across all settings of $k$ and voting schemes, although none of the results listed are able to beat the accuracy of baseline SNNs \cite{p._u._diehl_unsupervised_2015} of a comparable size.

\begin{table}
\centering
\captionsetup{justification=centering}
\caption{unshared weights, no \\connectivity, $n_\textrm{patches} = 50, s = 2$}
\label{unshared_no_connectivity}
\scriptsize
\begin{tabular}{|c|c|c|c|c|}
\hline
$k$ & \textit{all} & \textit{most-spiked} & \textit{top} \% & \textit{correlation} \\
\hline
8   & 80.21 $\pm$ 1.02\% & 76.69 $\pm$ 1.10\% & 73.83 $\pm$ 0.86\% & 79.57 $\pm$ 1.44\% \\
10  & 82.58 $\pm$ 1.33\% & 79.68 $\pm$ 0.78\% & 78.16 $\pm$ 0.98\% & 80.03 $\pm$ 1.30\% \\
12  & 83.56 $\pm$ 1.46\% & 81.18 $\pm$ 1.62\% & 80.50 $\pm$ 1.40\% & 81.20 $\pm$ 1.57\% \\
14  & \textbf{84.18} $\pm$ 1.55\% & 82.86 $\pm$ 1.54\% & 82.04 $\pm$ 1.35\% & 80.83 $\pm$ 1.07\% \\
16  & 80.02 $\pm$ 1.43\% & 79.39 $\pm$ 0.92\% & 76.82 $\pm$ 1.05\% & 78.75 $\pm$ 1.38\% \\
18  & 78.93 $\pm$ 1.20\% & 78.76 $\pm$ 1.18\% & 76.41 $\pm$ 1.28\% & 77.30 $\pm$ 1.23\% \\
20  & 75.82 $\pm$ 0.85\% & 76.22 $\pm$ 0.99\% & 74.38 $\pm$ 0.82\% & 73.88 $\pm$ 1.26\% \\
22  & 72.81 $\pm$ 1.06\% & 73.25 $\pm$ 1.05\% & 70.64 $\pm$ 1.04\% & 68.63 $\pm$ 1.61\% \\
24  & 67.53 $\pm$ 1.14\% & 66.51 $\pm$ 1.16\% & 65.65 $\pm$ 1.47\% & 64.67 $\pm$ 0.92\% \\
\hline
\end{tabular}
\end{table}

\begin{table}
\centering
\captionsetup{justification=centering}
\caption{unshared weights, pairwise \\connectivity, $n_\textrm{patches} = 50, s = 2$}
\label{unshared_pairwise_connectivity}
\scriptsize
\begin{tabular}{|c|c|c|c|c|}
\hline
$k$ & \textit{all} & \textit{most-spiked} & \textit{top} \% & \textit{correlation} \\
\hline
8   & 80.79 $\pm$ 1.27\% & 77.21 $\pm$ 1.55\% & 75.22 $\pm$ 1.56\% & 79.24 $\pm$ 1.26\% \\
10  & 82.95 $\pm$ 1.33\% & 79.44 $\pm$ 1.26\% & 78.39 $\pm$ 1.26\% & 80.37 $\pm$ 1.37\% \\
12  & 83.31 $\pm$ 1.37\% & 81.31 $\pm$ 1.21\% & 80.56 $\pm$ 1.50\% & 80.58 $\pm$ 1.06\% \\
14  & \textbf{84.23} $\pm$ 0.98\% & 83.19 $\pm$ 1.27\% & 82.02 $\pm$ 1.54\% & 81.17 $\pm$ 1.02\% \\
16  & 79.48 $\pm$ 1.45\% & 78.88 $\pm$ 1.52\% & 76.37 $\pm$ 0.92\% & 77.86 $\pm$ 1.52\% \\
18  & 78.77 $\pm$ 1.10\% & 78.77 $\pm$ 1.07\% & 77.33 $\pm$ 1.23\% & 77.66 $\pm$ 1.20\% \\
20  & 75.13 $\pm$ 1.45\% & 75.95 $\pm$ 1.11\% & 73.75 $\pm$ 1.41\% & 73.15 $\pm$ 1.46\% \\
22  & 73.29 $\pm$ 1.13\% & 73.10 $\pm$ 1.05\% & 71.90 $\pm$ 1.02\% & 70.90 $\pm$ 1.12\% \\
24  & 68.77 $\pm$ 1.52\% & 68.73 $\pm$ 1.17\% & 67.97 $\pm$ 1.27\% & 67.43 $\pm$ 1.35\% \\
\hline
\end{tabular}
\end{table}

\subsection{Convergence comparison}
\label{ssec:convergence}

The connectivity pattern from inhibitory to excitatory neurons creates virtual \textit{sub-networks} within each C-SNN; i.e., sub-populations of excitatory neurons compete to represent a particular region of input space. The kernel side length $k$ and stride $s$ determine the number of neurons per patch, while the number of patches parameter $n_\textrm{patches}$ describes how many neurons compete to represent features from each distinct visual field. This distributed method of network training allows for independent learning of many filters at once, granting the system an improvement in speed of feature learning and convergence to near-optimal classification accuracy.

\begin{figure} [ht]
  \centering
  \captionsetup{justification=centering}
  \includegraphics[width=0.475\textwidth, height=6cm]{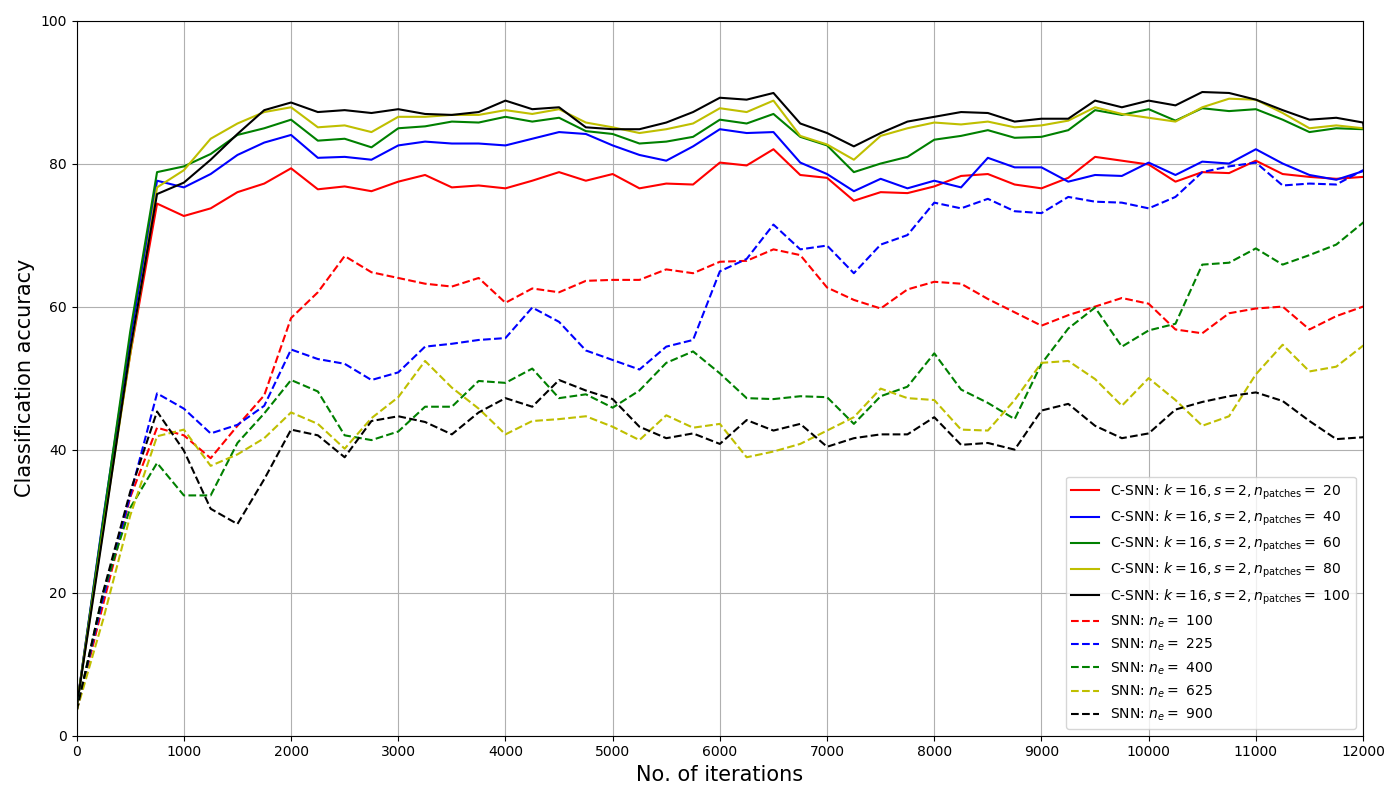}
  \caption{Smoothed estimated accuracy over baseline SNNs and C-SNNs training phase.}
  \label{fig:convergence}
\end{figure}

We show smoothed accuracy curves from single training runs over the first training 12,000 iterations for both baseline SNNs \cite{p._u._diehl_unsupervised_2015} and C-SNNs of various sizes in Figure \ref{fig:convergence}. Classification accuracy is evaluated every 250 iterations: neurons are assigned labels using network spiking activity and labels from the previous 250 training examples, and the subsequent 250 examples are classified according to the assigned labels and the \textit{all} voting strategy. C-SNNs tend to reach their optimal estimated accuracy after approximately 1,000 - 1,500 training examples, while larger SNNs have yet to catch up after all 12,000 iterations. Observe that there is no reason that networks should trend toward increasing for the entirety of training, as there is no supervision signal guiding parameter tuning or evaluation of the learned representation.

Since $k$ is fixed to 16 and $s$ is to 2, each network has $49 \times n_\textrm{patches}$ excitatory and inhibitory neurons, while the compared SNNs have at most 900 excitatory and inhibitory neurons. However, C-SNNs with this configuration learns only $16^2$ parameters per excitatory neuron compares to the baseline SNN's $28^2$, giving a total of $16^2 \times 49 \times n_\textrm{patches}$ parameters per network. While the number of neurons remains large compared to the baseline SNN, the number of parameters is manageable especially in the case of large networks.

\section{Conclusions and Future Work}
\label{sec:conclusions}

We have demonstrated a spiking neural network architecture capable of learning variously-sized features from image data. This distributed representation is possible due to the introduction of a windowed connectivity pattern between input and excitatory neuron populations, reminiscent of the successful convolutional neuron network layer in the deep learning literature. Excitatory and inhibitory connectivity create a competitive learning aspect in which either location-invariant or -dependent features can be learned faster than whole image prototypes at once.

Instead of a ``global'' learning rule which depends on information from a supervision signal, network parameters are tuned via a spike-timing-dependent plasticity rule. The relative timing and order of spikes of connected populations of neurons are utilized in combination with a inhibitory mechanism which encourages neurons to compete to represent different features. Unshared weights are modified at a rapid pace due to the introduction of inhibitory connectivity which encourages independent learning of different portions of input data.

The quality of learned image features is evaluated using neuron label assignments and several different voting strategies, in which recorded network activity is used to classify new input data. An important future direction is in developing a method of combining spiking activity from neurons with small filters into a meaningful description of the input data as a whole. Composing image features of multiple sizes may also yield useful global information to use for classification. Extending the C-SNN approach to more complex image datasets and to solve tasks (e.g., regression, clustering, and dimensionality reduction) may reveal general engineering principles behind spiking neural networks for machine learning.

Although state of the art classification performance is not attained, an interesting image data representation learning method is proposed. Moreover, this work represents an important step towards unifying ideas from deep learning and computing with spiking neural networks. We see many potential benefits from using spiking neural networks for machine learning tasks, namely computational scalability due to the independence of individual neuron simulations and local parameter updates, and the potential for efficient neuromorphic hardware implementation. On the other hand, utilizing more biologically motivated machine learning systems should allow researchers to make greater use of related concepts from the cognitive and brain sciences.

\section*{Acknowledgements}

This work has been supported in part by Defense Advanced Research Project Agency Grant, DARPA/MTO HR0011-16-l-0006 and by National Science Foundation Grant NSF-CRCNS-DMS-13-11165. M.R. has been supported in part by NKFIH grant 116769.

\bibliographystyle{IEEEtran}
\bibliography{zotero,add}

\begin{thebibliography}{10}
\providecommand{\url}[1]{#1}
\csname url@samestyle\endcsname
\providecommand{\newblock}{\relax}
\providecommand{\bibinfo}[2]{#2}
\providecommand{\BIBentrySTDinterwordspacing}{\spaceskip=0pt\relax}
\providecommand{\BIBentryALTinterwordstretchfactor}{4}
\providecommand{\BIBentryALTinterwordspacing}{\spaceskip=\fontdimen2\font plus
\BIBentryALTinterwordstretchfactor\fontdimen3\font minus
  \fontdimen4\font\relax}
\providecommand{\BIBforeignlanguage}[2]{{%
\expandafter\ifx\csname l@#1\endcsname\relax
\typeout{** WARNING: IEEEtran.bst: No hyphenation pattern has been}%
\typeout{** loaded for the language `#1'. Using the pattern for}%
\typeout{** the default language instead.}%
\else
\language=\csname l@#1\endcsname
\fi
#2}}
\providecommand{\BIBdecl}{\relax}
\BIBdecl

\bibitem{y._lecun_deep_2015}
{Y. LeCun, Y. Bengio, G. Hinton}, ``Deep learning,'' \emph{Nature}, no. 521,
  pp. 436--444, May 2015.

\bibitem{lee_neural_2015}
\BIBentryALTinterwordspacing
{Lee, S. W. and O'Doherty, J. P. and Shimojo, S.},
  ``\BIBforeignlanguage{en}{Neural {Computations} {Mediating} {One}-{Shot}
  {Learning} in the {Human} {Brain}},'' \emph{\BIBforeignlanguage{en}{PLOS
  Biology}}, vol.~13, no.~4, p. e1002137, Apr. 2015. [Online]. Available:
  \url{http://journals.plos.org/plosbiology/article?id=10.1371/journal.pbio.1002137}
\BIBentrySTDinterwordspacing

\bibitem{maass_lower_1996}
\BIBentryALTinterwordspacing
{Maass, W.}, ``Lower {Bounds} for the {Computational} {Power} of {Networks} of
  {Spiking} {Neurons},'' \emph{Neural Computation}, vol.~8, no.~1, pp. 1--40,
  Jan. 1996. [Online]. Available: \url{https://doi.org/10.1162/neco.1996.8.1.1}
\BIBentrySTDinterwordspacing

\bibitem{maass_networks_1997}
\BIBentryALTinterwordspacing
------, ``Networks of spiking neurons: {The} third generation of neural network
  models,'' \emph{Neural Networks}, vol.~10, no.~9, pp. 1659--1671, Dec. 1997.
  [Online]. Available:
  \url{http://www.sciencedirect.com/science/article/pii/S0893608097000117}
\BIBentrySTDinterwordspacing

\bibitem{p._u._diehl_unsupervised_2015}
{P. U. Diehl, M. Cook}, ``Unsupervised learning of digit recognition using
  spike-timing-dependent plasticity,'' \emph{Frontiers in Computational
  Neuroscience}, Aug. 2015.

\bibitem{y._lecun_gradient-based_1998}
{Y. LeCun, L. Bottou, Y. Bengio, and P., Haffner}, ``Gradient-based learning
  applied to document recognition.'' \emph{Proceedings of the IEEE}, vol.~86,
  no.~11, pp. 2278--2324, Nov. 1998.

\bibitem{liu_fast_2017}
\BIBentryALTinterwordspacing
{Liu, D. and Yue, S.}, ``Fast unsupervised learning for visual pattern
  recognition using spike timing dependent plasticity,'' \emph{Neurocomputing},
  vol. 249, pp. 212--224, Aug. 2017. [Online]. Available:
  \url{http://www.sciencedirect.com/science/article/pii/S0925231217306276}
\BIBentrySTDinterwordspacing

\bibitem{kheradpisheh_stdp-based_2016}
\BIBentryALTinterwordspacing
{Kheradpisheh, S. R. and Ganjtabesh, M. and Thorpe, S. J. and Masquelier, T.},
  ``{STDP}-based spiking deep convolutional neural networks for object
  recognition,'' \emph{arXiv:1611.01421 [cs]}, Nov. 2016, arXiv: 1611.01421.
  [Online]. Available: \url{http://arxiv.org/abs/1611.01421}
\BIBentrySTDinterwordspacing

\bibitem{diehl_fast-classifying_2015}
{Diehl, P. U. and Neil, D. and Binas, J. and Cook, M. and Liu, S. C. and
  Pfeiffer, M.}, ``Fast-classifying, high-accuracy spiking deep networks
  through weight and threshold balancing,'' in \emph{2015 {International}
  {Joint} {Conference} on {Neural} {Networks} ({IJCNN})}, Jul. 2015, pp. 1--8.

\bibitem{w._gerstner_spiking_2002}
{W. Gerstner, W. M. Kistler}, \emph{Spiking {Neuron} {Models}. {Single}
  {Neurons}, {Populations}, {Plasticity}}.\hskip 1em plus 0.5em minus
  0.4em\relax Cambridge University Press, 2002.

\bibitem{bi_synaptic_1998}
\BIBentryALTinterwordspacing
{Bi, G. and Poo, M.}, ``Synaptic {Modifications} in {Cultured} {Hippocampal}
  {Neurons}: {Dependence} on {Spike} {Timing}, {Synaptic} {Strength}, and
  {Postsynaptic} {Cell} {Type},'' \emph{Journal of Neuroscience}, vol.~18,
  no.~24, pp. 10\,464--10\,472, 1998. [Online]. Available:
  \url{http://www.jneurosci.org/content/18/24/10464}
\BIBentrySTDinterwordspacing

\bibitem{d._f._m._goodman_brian_2009}
{D. F. M. Goodman, R. Brette}, ``The {Brian} simulator,'' \emph{Frontiers in
  Computational Neuroscience}, Sep. 2009.

\bibitem{Rumelhart:1995:BBT:201784.201785}
\BIBentryALTinterwordspacing
D.~E. Rumelhart, R.~Durbin, R.~Golden, and Y.~Chauvin, ``Backpropagation,'' in
  \emph{Backpropagation}, {Chauvin, Y. and Rumelhart, D. E.}, Ed.\hskip 1em
  plus 0.5em minus 0.4em\relax Hillsdale, NJ, USA: L. Erlbaum Associates Inc.,
  1995, ch. Backpropagation: The Basic Theory, pp. 1--34. [Online]. Available:
  \url{http://dl.acm.org/citation.cfm?id=201784.201785}
\BIBentrySTDinterwordspacing

\end{thebibliography}

\end{document}